\documentclass[letterpaper,10pt,conference]{ieeeconf}
\IEEEoverridecommandlockouts
\usepackage{booktabs}
\usepackage{graphicx}
\usepackage{microtype}
\usepackage{amsmath,amssymb}
\usepackage{array}
\usepackage{url}
\usepackage[table]{xcolor}
\usepackage{wasysym}
\definecolor{specaccent}{HTML}{0B7A76}
\usepackage{hyperref}
\hypersetup{hidelinks}
\graphicspath{{figures/}}

\newcommand{\system}{OJOx}
\newcommand{\tY}{\CIRCLE}
\newcommand{\tP}{\LEFTcircle}
\newcommand{\tN}{\Circle}

\title{\system{}: Specification-Conditioned Demonstrations\\for Embodied AI in Construction}
\author{Mohamed Dawod\\[2pt]
\normalsize OJOx AI, London, United Kingdom\\[1pt]
{\tt\small m@ojox.ai}}

\begin{document}
\maketitle
\thispagestyle{empty}
\pagestyle{empty}

\begin{abstract}
Large-scale egocentric and whole-body human demonstrations are becoming a
primary source of data for embodied intelligence. They record what
people perceive and do, but rarely the external specification that gave an action
its purpose. In construction that omission is consequential: skilled work is
directed at project-specific configurations defined in a design model ---
configurations not yet present in the environment being observed. A mason's transferable
competence is not the geometry of one wall but the ability to realise a new
geometry from a specification. We introduce the
\emph{specification-conditioned demonstration}: a synchronised record of the
physical state a demonstrator perceives, the intended state supplied to them by
an external design, and the behaviour connecting the two. We present \system{},
a capture interface that realises this for construction --- delivering design
geometry to a headset, anchoring it in the physical workspace, rendering it into
a demonstrator's stereo passthrough view, and recording that view synchronously
with whole-body and hand motion. We report one fully instrumented
session --- a 33-component wall laid against a specification that changes while
the work proceeds --- and check the record against the physical scene through an
external camera registered independently of the capture. Recorded sessions remain
compatible with existing humanoid retargeting infrastructure and replay onto a
Unitree~G1 in simulation. The result is a data interface for testing whether
embodied policies can learn not merely to imitate demonstrated actions, but to
act toward specifications absent from their training experience.
\end{abstract}

\section{Introduction}
\label{sec:intro}

Embodied intelligence has entered a data-intensive phase. Egocentric video of
everyday human activity is now collected at the scale of thousands of hours
\cite{grauman2022ego4d,grauman2024egoexo4d}; hand and manipulation
demonstrations are gathered with purpose-built rigs \cite{chen2025arcap};
whole-body humanoid behaviour is captured through mocap-free teleoperation
\cite{ze2025twist2} and, increasingly, without any robot present at capture time
\cite{shi2026egohumanoid}. Between them these efforts are assembling a detailed
record of what people perceive and what they do.

\begin{figure*}[t]
  \centering
  \includegraphics[width=0.99\textwidth]{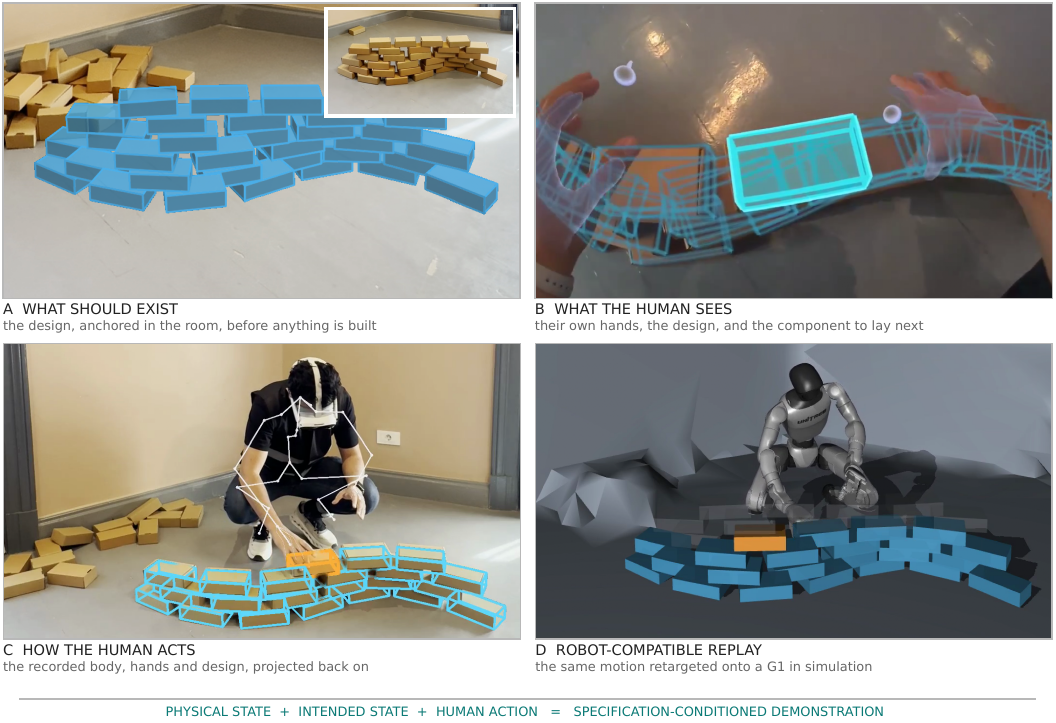}
  \caption{A specification-conditioned demonstration. \textbf{A} --- the design,
  anchored to the workspace, standing in the room before a single component has
  been laid, with the loose units piled at the left; the inset is the same camera
  and framing at the end of the work. \textbf{B} --- the demonstrator's own
  passthrough view: their hands, the workspace, and the component to lay next.
  \textbf{C} --- a static external camera that took no part in the capture, with
  the recorded body, hands and specification projected back onto the scene.
  \textbf{D} --- the same motion retargeted onto a Unitree~G1 in MuJoCo, from the
  same viewpoint. B, C and D are one moment. The robot is replaying recorded human
  motion in simulation: it is not executing the task, no object is simulated, and
  nothing is learned.}
  \label{fig:hero}
\end{figure*}

\begin{figure*}[t]
  \centering
  \includegraphics[width=0.99\textwidth]{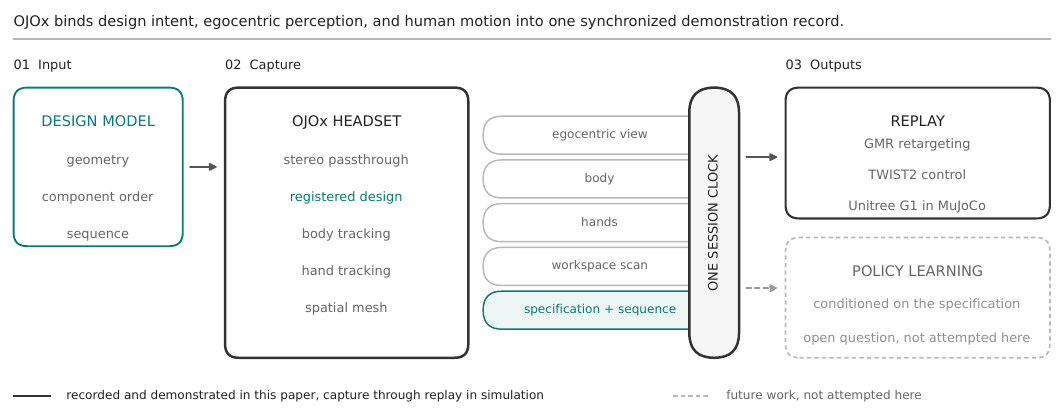}
  \caption{\system{} in three stages. A design model reaches the headset, which
  composites it into stereo passthrough, marks the component to lay next, and
  tracks the demonstrator against a spatial mesh of the room. Five streams leave
  the device and land on one clock; their convergence is the data object. Accent
  marks the specification wherever it appears. Replay in simulation is
  demonstrated here; conditioning a policy on the specification is not.}
  \label{fig:arch}
\end{figure*}

\begin{figure*}[t]
  \centering
  \includegraphics[width=0.99\textwidth]{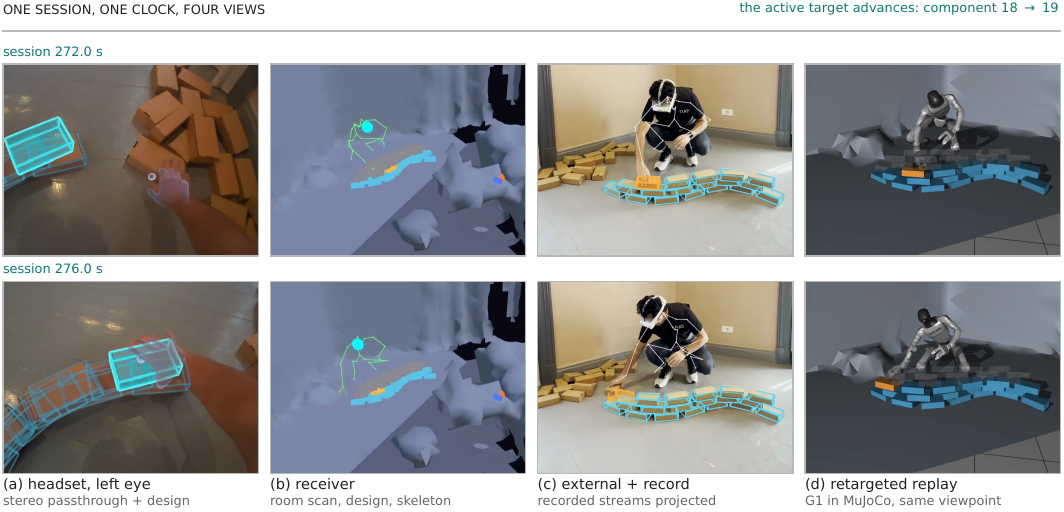}
  \caption{One session, one clock, four views --- at two instants four seconds
  apart, session 272.0\,s above and 276.0\,s below. Between them the demonstrator
  lays component~18 and the system advances the active target to component~19:
  the accent block jumps from the middle of the course to its far end, and the
  right arm follows it across. That single event is findable in every column,
  which is what a common clock buys and what a single instant could only assert.
  (a)~the headset's own rendered view; (b)~the receiver, holding the scanned room,
  the design with its active target, and the tracked skeleton; (c)~an external
  camera with the recorded streams projected in; (d)~the recorded motion
  retargeted onto a Unitree~G1 in MuJoCo, rendered from the same solved viewpoint
  as (c). (a)--(c) are recordings; (d) is offline replay in simulation.}
  \label{fig:fourviews}
\end{figure*}

There is an asymmetry in what that record contains. We have become adept at
capturing perception and action, and much less systematic about capturing the
\emph{specification} the person was working to satisfy. For a great deal of
manipulation this is a reasonable economy: when someone pours a cup or opens a
drawer, the objective can usually be read off the final frame, annotated
afterwards in language, or averaged across many examples of a recurring task. The
goal is latent in the data because the task is familiar and the world already
contains its answer.

Construction breaks that assumption. It is a \emph{specification-driven} domain:
what counts as correct is fixed by a design that is external to the scene and
frequently unique to one project. A worker positioning a panel is not
reproducing a canonical trajectory or reacting only to what is in front of them;
they are using general physical competence to realise a particular future state
that does not yet exist. The site tells the worker what exists. The
design tells them what should exist. The demonstration is how one becomes the
other. Two visually
near-identical motions can be right or wrong depending on a drawing, and no
amount of observing the scene resolves which.

We use \emph{intent} throughout in this external sense --- the specified physical
state a worker is acting to realise --- not in the psychological sense of
inferring a person's mental goals.

This distinction separates two learning problems that are easy to conflate. Much
of robot learning is imitation: \emph{here is what a person did; reproduce useful
behaviour}. Construction additionally demands something closer to
specification-following: \emph{here is a configuration you have never
encountered; use learned physical skill to realise it accurately in this
environment}. The transferable knowledge is therefore not a trajectory but a
relation --- between the current state, the desired state, and the skilled action
that closes the distance --- a mapping we make precise in
Section~\ref{sec:policy}.

Construction is a domain worth this trouble. It is economically vast, hazardous,
short of the skilled trades it depends on, and unusually suited to humanoids in
principle, since sites, tools and access routes are dimensioned around the human
body. Yet it remains among the least robotically penetrated physical domains, for
reasons that are well rehearsed: unstructured and continually changing
environments, mobility coupled with dexterity, tight tolerances, and
specifications that differ from project to project \cite{uthai2025roadmap}.
Industry analysis foregrounds those site conditions and unit cost as the
principal deployment barriers, while noting that humanoid skill acquisition
presupposes sufficient demonstration data \cite{mckinsey2025humanoid}. That prerequisite is
worth taking seriously: existing egocentric corpora are dominated by domestic and
tabletop activity, and robot density is reported only for manufacturing, with no
construction equivalent published \cite{ifr2025density}. Skilled trade work is
therefore worth capturing while that expertise is abundant.

Our proposal is to stop treating the specification as something to be recovered
after the fact, and to place it in the demonstrator's perception while they work.
Construction makes this unusually tractable, because it is one of the few
physical domains in which a machine-readable description of the intended world
exists \emph{before} that world does: the design model. Rendering that model into
the worker's own view is an interface construction AR has already validated for
guidance: controlled studies with practitioners and manual assemblers report
reduced task time and error relative to 2D drawings, though registration accuracy
remains limiting for tight tolerances
\cite{chalhoub2019ar,kwiatek2019ar,yang2019ar,mitterberger2020augmented}. The
step we take is to record that guidance rather than only display it, so the data
contains the specification alongside the scene and the behaviour
(Fig.~\ref{fig:hero}).

We call the result a \emph{specification-conditioned demonstration}: a
synchronised record of the perceived physical state, the externally supplied
intended state, and the human behaviour connecting them. The point is not an
extra metadata column. Because the specification is visible \emph{during} the
action, it is available to shape approach, alignment, orientation, correction and
recovery --- couplings that a goal inferred from the final state cannot
reconstruct, and that remain directly observable even in partial or abandoned
attempts.

\system{} is a working implementation of this idea for construction
(Fig.~\ref{fig:arch}), built by adapting infrastructure developed for humanoid
teleoperation into a
\emph{robot-free} capture system: the worker demonstrates in their own body, no
humanoid present, and retargeting happens offline. This removes the
requirement to station a humanoid at the collection site, an important
prerequisite for future scaling of trade-data collection.

\paragraph*{Contributions}
\begin{enumerate}
  \item We identify the missing specification as a distinct data problem for
        embodied construction: the state an action is directed at is often absent
        from the observable scene and available only from the design, so
        conventional demonstrations under-determine the task.
  \item We define the \emph{specification-conditioned demonstration} as the data
        object that repairs this, binding observation, intended state and
        behaviour at collection time rather than by later annotation.
  \item We present \system{}, a capture architecture that delivers design
        geometry to a headset, anchors it in a worker's workspace, and records
        the resulting egocentric view synchronously with whole-body motion, hand
        motion, and a machine-readable specification record.
  \item We validate that data on one fully instrumented session --- a
        33-component assembly whose specification changes while the work
        proceeds --- checking the record against the physical scene through an
        independently registered external camera, and replaying the recorded
        motion through GMR/TWIST2 onto a 29-DoF Unitree~G1 in simulation.
\end{enumerate}

\section{Related Work}
\label{sec:related}

\system{} draws on five lines of work. Existing systems provide these
ingredients separately.

\paragraph{Human demonstration data for embodied intelligence}
Ego4D and Ego-Exo4D established egocentric capture of everyday activity at
scale \cite{grauman2022ego4d,grauman2024egoexo4d}. HoloAssist records
mixed-reality-mediated collaborative tasks with head, hand and gaze streams,
where guidance reaches the performer verbally from a remote instructor
\cite{wang2023holoassist}. RH20T pairs robot episodes with human demonstration
video \cite{fang2023rh20t}, and IndustReal captures procedural industrial
assembly, releasing CAD models of the parts while delivering the instructions to
workers on paper \cite{schoonbeek2024industreal}. These datasets do not generally embed a
design-derived, spatially registered intended state in the demonstrator's
egocentric observation.

\paragraph{Whole-body capture and humanoid retargeting}
TWIST2 collects whole-body humanoid demonstrations through mocap-free
teleoperation, the operator driving the robot while seeing through its cameras
\cite{ze2025twist2}; GMR supplies the retargeting from human to humanoid
kinematics \cite{araujo2025gmr}; Unitree's stack builds on the same headset
family \cite{unitree2026xrteleoperate}. A newer strand removes the robot from
capture altogether, aligning egocentric human demonstrations with a smaller
robot corpus \cite{shi2026egohumanoid}. \system{} adopts this line's tracking
protocol and retargeting stack wholesale, and inherits its robot-free property;
what they do not provide is the design-derived spatial specification that the
demonstrator is acting to realise.

\paragraph{Goal-conditioned policies and rendered goals}
Policies are routinely conditioned on visually expressed goals --- goal images,
hand-drawn sketches \cite{sundaresan2024rtsketch}, or targets rendered directly
into the observation \cite{shridhar2024genima}. Of these, AnySlot is the
nearest in mechanism:
a spatial marker denoting the intended placement is rendered into the policy's
camera streams and kept view-consistent, so the goal is carried in the
observation rather than in a pose vector \cite{hu2026anyslot}. The goal there is
produced generatively from language and rendered for the \emph{robot} rather
than shown to a human demonstrator. The broader pattern in this literature is that
goal signals are manufactured after collection --- by hindsight relabelling or
by user input at test time --- rather than being present, and shaping behaviour,
while the demonstration is performed.

\paragraph{Mixed reality in demonstration collection}
Microsoft's SIGMA renders procedural task guidance as spatially anchored
holograms and captures them within the egocentric stream, released as the
SigmaCollab dataset \cite{bohus2024sigma,bohus2025sigmacollab}; the guidance is
a hand-authored recipe rather than a design model, and no whole-body or
humanoid-retargetable motion is recorded. In AR-assisted robot data collection
the overlay generally serves the operator rather than the task: ARCap shows a
robot twin and feasibility feedback to improve demonstration quality
\cite{chen2025arcap}, ARMimic likewise records a virtual robot in the captured
view \cite{walia2025armimic}, and EgoGuide renders online data-coverage cues
during robot-free collection that are deliberately excluded from the recorded
stream and from policy input \cite{xu2026egoguide}. These systems establish that
in-headset guidance during capture is practical; what they render is embodiment
or data-quality information, not the intended state of the work.

\paragraph{Design information in construction robotics}
Augmented Bricklaying rendered design-derived placement targets into masons'
views through object-aware registration, guiding the assembly of a
13{,}596-brick facade \cite{mitterberger2020augmented} --- the guidance half of
our premise, executed at building scale, but recording no worker motion. BIM has
been used as a perception prior for on-site assembly \cite{dawod2019bim} and as
a spatial prior for construction robot navigation \cite{kim2025bimnav}; in both
the model informs planning rather than appearing in a demonstration record.
Construction skill learning itself is beginning to attract attention: recent
work benchmarks vision-language-action policies against hierarchical RL on
multi-stage panel installation using teleoperated demonstrations
\cite{hu2025constructionskill}, and humanoids have been taught construction
skills from third-person observation of workers
\cite{liu2026humanoidconstruction} --- third-person observation, with no
specification channel in the recording.

\paragraph{The intersection}
Stated precisely: we are not aware of prior work that simultaneously records a
\emph{design-derived} intended state that is spatially registered to the
workspace, \emph{shown to a human demonstrator during the demonstration} and
thereby captured inside the egocentric stream, \emph{time-synchronised with
whole-body and hand motion suitable for humanoid retargeting}. Prior work
renders spatial goals into a \emph{robot's} observations from generative or
language sources \cite{hu2026anyslot}, shows in-headset guidance that is
deliberately kept out of the recorded stream \cite{xu2026egoguide}, records
authored procedural holograms without whole-body capture
\cite{bohus2025sigmacollab}, or uses design models for planning and navigation
rather than as a recorded target state \cite{dawod2019bim,kim2025bimnav}. We
claim the conjunction, not any single ingredient.

\section{OJOx: Specification-Conditioned Capture}
\label{sec:system}

Augmented reality is the mechanism here, not the contribution. What the system
produces is a data interface (Fig.~\ref{fig:arch}): a way of getting an externally specified intended
state into a demonstration at the moment the demonstration happens.

\subsection{Design specification}
The specification is design geometry --- in construction practice originating in
BIM/CAD tooling --- packaged with its task data and published to cloud storage,
from which the headset retrieves it over HTTP and caches it on the device. The
same loader accepts a package side-loaded onto the device, and the session
reported in Section~\ref{sec:characterization} was captured that way: cloud
delivery is the deployment path the architecture is built around and is
implemented in the client, but it is not the path that session exercised. We
distinguish the two throughout.

A package carries geometry, component identity, placement pose, and
assembly-sequence state. Components are separately named nodes inside the
delivered model, and their order is the intended assembly order --- for the
reported session the geometry was authored parametrically and exported course by
course, so that ordering came from the design tool rather than from the capture
system. Sequence transitions --- revealing the next component of an ordered
assembly --- update the active visual specification while the demonstration is
under way and are propagated through the same specification stream. Richer BIM
semantics (tolerances, dependency graphs, construction-process logic) are
representable in the same channel but are not used by the present system; we
return to this in Section~\ref{sec:agenda}.

\subsection{Design-to-workspace anchoring}
The delivered geometry then has to be brought into correspondence with the
physical workspace. In the capture system as it stands the demonstrator does this
directly: the device reconstructs a spatial mesh of the room, the model is
positioned by raycasting against that mesh, and once confirmed it stays where it
was put, world-anchored by the headset's own tracking. Two properties matter for
the data. The anchor is physical rather than view-relative --- the model rests on
the scanned floor and walls, and does not drift with the head. And the resulting
placement is not left implicit in the imagery: the model's identity, 6-DoF pose
and scale are written into the session record, so where the specification sat in
the workspace is recoverable from the data and not only visible in the video.

A marker-referenced path would remove the operator from that step: markers placed
at reference points with known design-frame coordinates, and a table in the
delivered package associating each code with its pose in that frame, so resolving
one marker fixes the design-to-workspace transform outright. This is implemented
in the platform's handheld capture client. On the headset the corresponding
component exists in the codebase but is not attached to the runtime, and the
session reported in Section~\ref{sec:characterization} did not use it: that
session was anchored by the operator against the device's spatial reconstruction,
as described above. We report no accuracy figure for either path
(Section~\ref{sec:agenda}).

\subsection{A shared visual specification}
The demonstrator sees the physical workspace through stereo passthrough with the
intended state composited into it, and works normally. Nothing about the task
arrives through a separate instruction channel: the rendered geometry, seen in
place, \emph{is} the specification. This is the same interface construction AR
has used for human guidance \cite{mitterberger2020augmented}; the difference is
that here it is also recorded.

\subsection{Whole-body, hand and egocentric capture}
Capture reuses the established headset recipe: head pose, a 24-joint body
skeleton via two motion trackers, 26 joints per hand, and controller state,
streamed to a macOS receiver. Nothing pins the sample rate in software; it
follows the headset's own frame rate, so we report what a session achieved
rather than a specification. The receiver records the demonstrator's rendered
egocentric view --- passthrough scene plus composited specification --- at
4320$\times$2160 and 30\,fps. That stream is the load-bearing property of the
data: it contains what the worker saw physically \emph{and} what the task
required visually, in the same pixels. A scan pass before recording streams the
device's reconstruction of the room to the receiver, which writes it into the
session as a single mesh, so the workspace the demonstration happened in is part
of the record.

\subsection{Synchronisation and session representation}
All streams carry a common session clock. A session is a self-contained unit ---
tracking stream, rendered-view video, metadata --- and additionally holds a
machine-readable specification record: what the model is, where it sits in the
workspace, and how far through the assembly the demonstrator has been taken. The
specification is therefore present twice over, visually and symbolically, which
is what lets Section~\ref{sec:policy} compare the two from one dataset.

The two forms are stored differently, and deliberately. The visual one is a
continuous stream; the symbolic one is sparse, written when the intended state
changes rather than copied into every frame, with the head and wrist positions at
that instant alongside it. Semantic change is thereby preserved without being
duplicated forty thousand times, and stays alignable to the continuous streams.

The same record carries candidate pick and place events. These come from the hand
stream alone --- an aperture threshold with a wrist-speed gate --- so they mark
when a hand closed and opened, not when an object was grasped and released. No
object is tracked and no hand-to-object offset is estimated. They are context
around each transition, not object-level ground truth, and
Section~\ref{sec:characterization} shows what that costs.

\subsection{Humanoid compatibility}
\label{sec:retarget}
Sessions replay offline through the established humanoid pipeline: frames are
retargeted by GMR (\texttt{xrobot} $\rightarrow$ \texttt{unitree\_g1})
\cite{araujo2025gmr} and driven through the TWIST2 control stack
\cite{ze2025twist2} onto a 29-DoF Unitree~G1 in MuJoCo. The one task-specific
adaptation is a retargeting configuration rather than a change to the stack:
wrist position is weighted an order of magnitude above the other tracked links,
because in this task the accuracy that matters is at the hands. Recorded finger
joints drive the robot's hands through a direct mapping of bone-to-bone flexion
angles; the three-fingered hand model has no counterpart for ten of the
twenty-six tracked joints per hand, and those are unused. We use GMR/TWIST2 as
the downstream embodiment mapping; evaluating retargeting fidelity is outside the
scope of this work. The architectural point is that the specification layer
rides the existing session format, so no robot need be present at capture time
and the added specification signal requires no modification to the downstream
retargeting interface.

\section{The Specification-Conditioned Demonstration}
\label{sec:dataobject}

Write one session as
\begin{equation}
  \mathcal{D} = \{(o_t,\, g_t,\, h_t)\}_{t=1}^{T},
  \label{eq:dataobject}
\end{equation}
with $o_t$ the observation of the physical state, $g_t$ the externally supplied
intended state, and $h_t$ the demonstrator's whole-body and hand configuration.
In \system{}, $o_t$ and the visual form of $g_t$ arrive fused: the specification
is composited into the same pixels the demonstrator perceives. The symbolic
record provides $g_t$ a second time, as component identity, 6-DoF pose, and
current assembly-sequence state on the shared session clock. $h_t$ comprises the
24-joint body, dual 26-joint hands, head and controller state on that same
clock.

Construction specifications are not only spatial but temporal: as work
progresses, the relevant intended state changes. Accordingly $g_t$ need not be
static. In the current system it may advance during a demonstration as the
assembly sequence steps --- successive components are revealed, and the active
design state is propagated through the same specification stream that carries the
placed geometry. The record can therefore hold not only what should exist, but
what should exist \emph{next}.

What separates $\mathcal{D}$ from a conventional demonstration is not an
additional field but \emph{when} the binding happens. Goal and action are
synchronised at collection time, not associated afterwards. Three consequences
follow.

First, the specification is available to shape the behaviour as it is produced
--- approach direction, body placement, alignment, orientation, correction,
sequencing, when to stop, when to retry. Hindsight relabelling, which infers a
goal from a demonstration's final frame, cannot reconstruct that coupling; it
can only observe its result.

Second, the record remains meaningful for attempts that fail or stop partway. A
demonstration abandoned halfway has no useful final state to relabel from, but it
does have a specification and a trajectory of an expert not reaching it --- the
data a system would need in order to study correction. The same applies within a
session: when a demonstrator steps the assembly state backwards, that reversal is
in the record as a change in $g_t$, not lost in a summary of what was eventually
built (Sec.~\ref{sec:session}).

Third, expressing the goal visually rather than as a task class keeps it valid
as the component vocabulary and site layout change. This suggests the rendered
specification as a candidate \emph{shared representation}: the worker reads
scene-plus-intent and acts; a policy could in principle read scene-plus-intent
and act (Fig.~\ref{fig:shared}). We advance that symmetry as a hypothesis the
data makes testable, not as a result --- and note that \system{} records the
symbolic form too, so the comparison need not presuppose the answer.

\section{System Validation}
\label{sec:characterization}

\begin{table*}[t]
  \caption{\system{} in the landscape}
  \label{tab:landscape}
  \centering
  \footnotesize
  \setlength{\tabcolsep}{1.6pt}
  \begin{tabular}{lccccccccc}
    \toprule
    & \rotatebox{60}{Whole-body} & \rotatebox{60}{Hands} & \rotatebox{60}{Egocentric video}
    & \rotatebox{60}{\shortstack[l]{Humanoid\\retargeting}} & \rotatebox{60}{\shortstack[l]{Explicit goal\\recorded}}
    & \rotatebox{60}{\shortstack[l]{Goal in ego\\stream}} & \rotatebox{60}{\shortstack[l]{Goal world-\\registered}}
    & \rotatebox{60}{\shortstack[l]{Goal from\\design model}} & \rotatebox{60}{\shortstack[l]{Seen by human\\demonstrator}} \\
    \midrule
    Teleoperation (TWIST2, GMR) \cite{ze2025twist2,araujo2025gmr} & \tY & \tY & \tY & \tY & \tN & \tN & \tN & \tN & \tN \\
    Robot-free humanoid capture \cite{shi2026egohumanoid} & \tY & \tY & \tY & \tY & \tN & \tN & \tN & \tN & \tN \\
    Egocentric datasets \cite{grauman2022ego4d,grauman2024egoexo4d,wang2023holoassist} & \tP & \tP & \tY & \tN & \tP & \tN & \tN & \tN & \tP \\
    MR task assistance \cite{bohus2024sigma,bohus2025sigmacollab} & \tN & \tY & \tY & \tN & \tY & \tY & \tP & \tN & \tY \\
    AR-assisted collection \cite{walia2025armimic,chen2025arcap,xu2026egoguide} & \tN & \tY & \tP & \tP & \tN & \tP & \tP & \tN & \tY \\
    Rendered goals for policies \cite{shridhar2024genima,hu2026anyslot} & \tN & \tN & \tP & \tN & \tY & \tY & \tP & \tN & \tN \\
    AR construction guidance \cite{mitterberger2020augmented} & \tN & \tN & \tN & \tN & n/a & n/a & \tY & \tY & \tY \\
    \midrule
    \rowcolor{specaccent!8}
    \textbf{\system{} (ours)} & \tY & \tY & \tY & \tY$^{\ddagger}$ & \tY & \tY & \tY$^{\dagger}$ & \tY & \tY \\
    \bottomrule
  \end{tabular}

  \vspace{3pt}
  \parbox{\textwidth}{\footnotesize \CIRCLE~yes \quad \LEFTcircle~partial \quad
  \Circle~no \quad n/a~not applicable.\; Rows group representative systems.
  Several columns lie outside a given system's purpose rather than marking a
  deficiency, and where no goal representation is recorded at all the four goal
  columns are negative by entailment. $^{\dagger}$The specification is anchored to
  the device's own spatial reconstruction of the workspace and its 6-DoF pose in
  that frame is recorded; the marker-referenced bridge to a design frame is
  implemented in the platform's handheld client and was not used on the headset
  (Sec.~\ref{sec:system}-B). $^{\ddagger}$In MuJoCo simulation.}
\end{table*}

This section establishes that the capture interface exists and operates end to
end. It is not a policy-learning evaluation, and it does not measure task
performance, registration accuracy, or worker productivity.

The evidence comes from two generations of recording. The earliest pilot sessions
predate the machine-readable specification record and the assembly-sequence path
and exercise neither; they evidence specification-conditioned capture and
downstream compatibility, and nothing about the symbolic record or the sequence.
Everything below is from the instrumented session of
Table~\ref{tab:session}. Neither generation exercises marker-referenced
anchoring, which is implemented in the handheld client but not wired into the
headset runtime.

\subsection{One instrumented session}
\label{sec:session}

\begin{table}[t]
  \caption{The instrumented session. Every figure here is recomputed from the
  session's own files by the verification script accompanying this paper.}
  \label{tab:session}
  \centering
  \footnotesize
  \setlength{\tabcolsep}{4pt}
  \begin{tabular}{@{}p{0.40\columnwidth}p{0.54\columnwidth}@{}}
    \toprule
    Duration & 528.6\,s (8\,min 49\,s), indoors, one demonstrator \\
    Tracking samples & 45{,}000; 85.0\,Hz mean, inter-sample median 11.4\,ms
                       (5th--95th pct.\ 8.3--16.2\,ms) \\
    Body, hands & 24 body joints; 26 joints per hand, both hands \\
    Hand tracking available & 87.0\,\% left, 85.6\,\% right, 83.2\,\% both \\
    Rendered egocentric view & 4320$\times$2160, 30\,fps \\
    Room scan & 3{,}784 vertices, 5{,}235 triangles \\
    Specification & 33 components, 6 courses, 180$\times$95$\times$50\,mm each \\
    Specification records & 553 in the tracking stream ($1.05\,\mathrm{s}^{-1}$);
                            46 assembly events \\
    Presentation & whole model 0.8--62.5\,s, then sequential targeting \\
    Sequence transitions & 33 forward, median 12.3\,s apart (3.2--47.5\,s);
                           6 steps backwards \\
    Grasp events & 67 pick, 40 place --- hand-inferred, not object-level;
                    two hold durations implausible \\
    \bottomrule
  \end{tabular}
\end{table}

\begin{figure*}[t]
  \centering
  \includegraphics[width=0.99\textwidth]{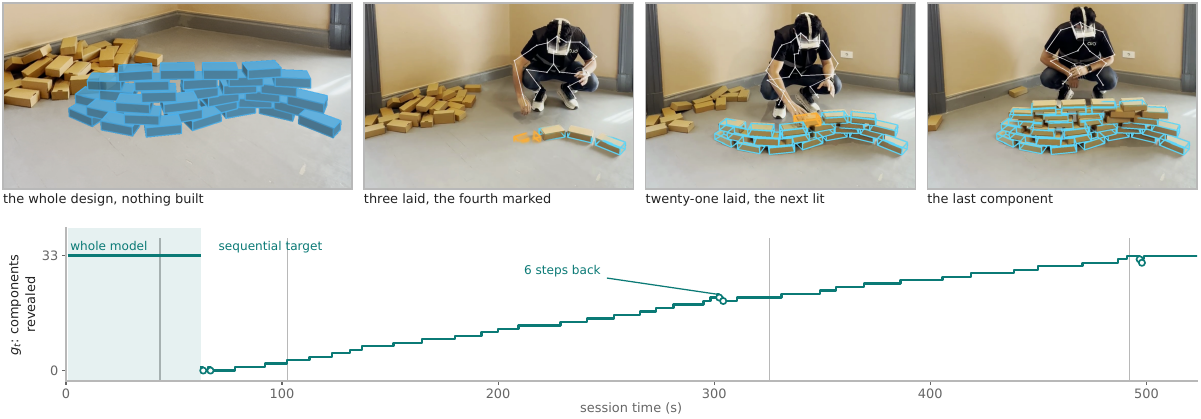}
  \caption{The target state changes while the human acts. Below: the
  assembly-sequence state $g_t$ over the session clock, read from the session's
  own assembly-event record at its recorded timestamps. The complete
  33-component design is displayed for the first minute; the system then switches
  to sequential targeting and reveals components one at a time. Open circles mark
  the six transitions that step \emph{backwards}. Above: four recorded frames at
  the marked instants. A goal recovered from the finished wall would contain none
  of this.}
  \label{fig:spectime}
\end{figure*}

An operator wearing the headset laid a curved, 33-component wall of cardboard
units against the delivered design, working from a pile of loose units to one
side, over about nine minutes. Table~\ref{tab:session} is the record, and what follows
works through it: the specification was present, it changed while the work went
on, it was synchronised with the motion that responded to it, the record can be
checked against the physical scene, and the motion still replays onto a
humanoid.

\paragraph{The specification was present, in two forms}
The demonstrator worked while perceiving the rendered design: the recorded
egocentric view holds the physical workspace, their own hands and the
design-derived target in the same pixels (Fig.~\ref{fig:hero}B). The claim is
perceptual availability, and it is what the recorded video shows; no task-scoring
protocol was run. Alongside it the record carries the specification symbolically: a heartbeat about
once a second, and an event written at each instant it changed, carrying the
model's identity, its pose in the workspace and the sequence state. The
comparison proposed in Section~\ref{sec:policy} needs both forms from the same
session, and this one has both.

\paragraph{It changed while the work went on}
The complete assembly was displayed for the first minute. The system then
switched to sequential targeting and named one component at a time, a new target
roughly every twelve seconds, the last of them not until the eighth minute.
Six of those transitions step \emph{backwards} --- in one, the target retreats two
components before advancing again. Fig.~\ref{fig:spectime} is the trace. A goal
recovered from the finished wall would contain none of it --- not the order, not
the pacing, and not the reversals, which are the only sign in the data of the
demonstrator judging that something had gone wrong.

\paragraph{It was synchronised with the motion that responded to it}
All of it shares one clock: the rendered egocentric view, a 24-joint body
skeleton, both hands at 26 joints each, head pose and controller state. Tracking
held at 85\,Hz with the specification rendered and its symbolic record being
written, inside the band the capture recipe delivers without them; we did not
measure the layer's computational cost separately. Optical hand tracking was
available for 83\,\% of frames on both hands at once. Fig.~\ref{fig:fourviews} is the alignment seen from outside: one
target advance, found in four independently produced recordings four seconds
apart.

\paragraph{What the record does not contain}
It does not contain the material. Candidate pick and place events are inferred
from hand aperture and wrist motion, which makes them useful local action context
and nothing more: no object is tracked, so they mark when a hand closed rather
than what it closed on, and against 33 components the detector plainly
over-triggers (Table~\ref{tab:session}).

\subsection{The record against the physical scene}

\begin{figure*}[t]
  \centering
  \includegraphics[width=0.99\textwidth]{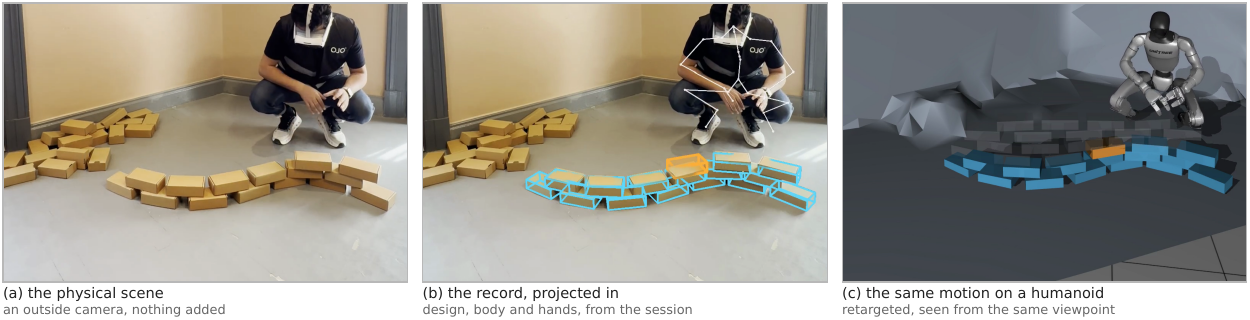}
  \caption{The record checked against the physical scene, at one moment.
  (a)~an external camera that took no part in the capture, unmodified.
  (b)~the same frame with the recorded specification, body and hands projected
  into it: every drawn element is read back from the session, and the only thing
  the video supplies is the viewpoint. (c)~the same recorded motion retargeted
  onto a Unitree~G1 in MuJoCo, seen from that same viewpoint, which is what
  allows (b) and (c) to be compared. The robot is replaying recorded human
  motion in simulation: it is not executing the task, no object is simulated,
  and nothing is learned.}
  \label{fig:session}
\end{figure*}

Nothing so far establishes that the recorded specification and the recorded body
agree with the physical world: a self-consistent record can be uniformly wrong.
The session was therefore filmed from a static external camera that took no part
in the capture. Everything drawn on that footage in Fig.~\ref{fig:session} is read
back from the session and projected through a single viewpoint. The design lands
on the bricks; that is the check, and it is a visual one.

The viewpoint itself was solved independently of the \system{} capture, from
manually identified correspondences between design-frame component corners and
image pixels, and fits them to a median reprojection error of 1.4\,px. An
independently recorded head trajectory, which took no part in that solve,
reprojects through the same camera at a median of 17.2\,px over 980 samples.
These values validate the external reprojection used for visualisation. They are
not a measurement of \system{}'s design-to-workspace registration, which remains
unquantified (Sec.~\ref{sec:agenda}).

What the projection does expose is a gap, and it is not a registration error.
The design is six courses of single-unit thickness; what was built follows the
same curve more thickly and more roughly. Silhouette agreement between the two is
0.42, and no rigid re-placement of the design rescues it --- the best one reaches
0.46. That ceiling is set by the shape difference, not by the camera, and it is
precisely the distance between what was specified and what was made. It is the kind of
quantity this data exists to expose, and one this session can only bound:
segmenting the touching cardboard units was not reliable enough to support a
per-component error, and that measurement remains outstanding.

The same view also bounds what the body stream measures. Asked whether each
recorded joint projects inside the operator's own silhouette, the head-and-torso
chain lands inside nine frames in ten; the elbows manage fewer than five. The
headset solves the body from the head and the hands, so limbs it cannot see are
inferred rather than observed, and elbow and lower-limb positions should be read
as estimates. The hands are the exception: they are optically tracked rather than
inferred.

\subsection{Replay onto a humanoid}

A recorded session replays unmodified through the GMR/TWIST2 path onto the 29-DoF
Unitree~G1 in MuJoCo (Fig.~\ref{fig:hero}D, Fig.~\ref{fig:fourviews}d). No change
to the retargeting stack was required: the specification layer rides the existing
session format, so compatibility holds by construction, and it is confirmed in
practice. The task-specific IK weighting of Section~\ref{sec:retarget} is a
configuration that stack consumes, not a modification of it.

All of this is simulation. The robot replays recorded human motion: it does not
perform the task, no object is simulated, no policy is trained, and nothing about
the demonstration is learned. What this section establishes is that the data
exists, that it holds what we claim it holds, and that it survives the mapping
onto a humanoid's kinematics.

\section{Toward Specification-Conditioned Embodied AI}
\label{sec:policy}

\begin{figure}[t]
  \centering
  \includegraphics[width=\linewidth]{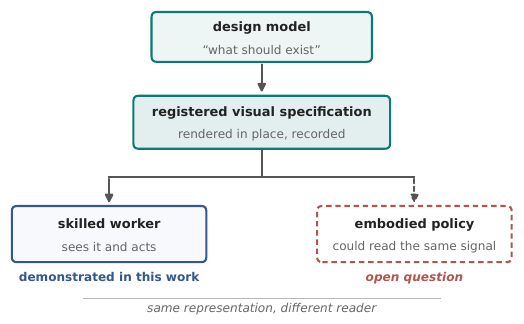}
  \caption{One spatial representation, two readers. The design model states what
  should exist; registering it to the workspace turns it into a specification a
  worker can see and act on, and --- prospectively --- a signal a policy could
  consume in the same form. The dashed branch is not a result: whether it holds
  is the open question this data is built to test.}
  \label{fig:shared}
\end{figure}

A conventional imitation dataset supports one mapping, $o_t \rightarrow a_t$:
act from what you see. A specification-conditioned demonstration makes a
different one learnable in principle,
\begin{equation}
  (o_t,\; g_t) \;\rightarrow\; a_t,
  \label{eq:policy}
\end{equation}
which reads: given what exists and a statement of what should exist, how should
I act? The shift it describes is from \emph{imitate this action} to
\emph{realise this specification}. None of what follows is demonstrated here; it
is the programme the interface exists to enable.

Without an explicit $g_t$, the desired configuration must be inferred implicitly
from the observation or task context. Conditioning on $g_t$ instead exposes that
variation directly to the policy, making generalisation across held-out
specifications --- a bond pattern the policy has never laid --- an explicit
empirical question. Whether that generalisation materialises is untested, and it
is the question we think the field should be able to ask of construction data.

\paragraph{The experiment this enables}
Train a visuomotor policy on \system{} demonstrations under three conditionings:
(A) observation only, (B) observation plus the symbolic specification, (C)
observation plus the rendered visual specification. Evaluate on component
selection, target position, orientation and execution --- and, critically, on
held-out specifications. Because \system{} records both forms of $g_t$,
conditions B and C come from the same sessions, so the comparison is a property
of the dataset rather than of separate collection efforts.

That design is what keeps the shared-representation idea honest. The symmetry in
Fig.~\ref{fig:shared} is appealing --- the same rendered geometry read by a
worker and by a policy --- but appeal is not evidence. It is entirely possible
that a symbolic pose vector is the better conditioning signal and the rendered
form merely convenient for humans. The value of recording both is that the
question is decidable.

\section{Discussion and Research Agenda}
\label{sec:agenda}

\paragraph{From prototype to site}
Every session here was captured indoors. They establish the data primitive, not
robustness to an active site: outdoor lighting, dust, occlusion, moving workers
and the ordinary disorder of construction all remain untested.

\paragraph{Anchoring accuracy}
The specification is anchored to the device's own reconstruction of the room, and
its absolute accuracy --- placement error, drift under head motion, end-to-end
design-to-site error --- is unquantified. The reprojection figures in
Section~\ref{sec:characterization} validate the external camera used to visualise
the record; they say nothing about this. The marker-referenced path that would tie
the anchor to a design frame exists in the handheld client but not in the headset
runtime, so on the headset the anchor is currently set by the operator. For a
domain defined by tolerance this is the most consequential open measurement, and
it bounds any future claim about precision.

\paragraph{What the body stream measures}
The headset solves the body from the head and the hands. Limbs it cannot see are
inferred, and the external check in Section~\ref{sec:characterization} finds them
outside the operator's own silhouette more often than not. Any use of this data
should treat elbow and lower-limb positions as estimates. The hands are the
strongest part of the stream, being optically tracked, but they too were
unavailable for part of the session.

\paragraph{No object state}
The system records the specification and the demonstrator; it does not record the
material. No object is tracked and no hand-to-object offset is estimated, so the
inferred pick and place events say when a hand closed, not what it closed on. What
was actually built is unmeasured too: the shape gap reported in
Section~\ref{sec:characterization} bounds it, but no per-component as-built error
is established, and a metric reconstruction of the finished work is the obvious
next measurement.

\paragraph{Delivery path}
Cloud delivery of the specification is implemented in the headset client and is
the deployment path the architecture assumes, but the session reported here was
captured with the package side-loaded onto the device. The claim we make is for
the interface and the recorded data, not for a fielded delivery service.

\paragraph{Sequence semantics}
\system{} supports an evolving ordered assembly state, and sequence transitions
are recorded alongside the demonstration. The sequencer, however, represents
reveal and progression through an ordered component list rather than a
construction process model. Tolerances, dependency and precedence constraints,
automatic inference of which components are physically complete, and sequence
planning are all absent; the order is authored, not derived. A specification that
carries these is the natural extension.

\paragraph{Data scale and diversity}
This paper validates an interface, not a dataset: one session, one demonstrator,
one design, indoors. Scaling means more sessions, but also more demonstrators ---
an interface that captures one worker's idiom is not yet capturing a trade. Worker diversity, task diversity, and site diversity are all
prerequisites for the learning question to be answerable.

\paragraph{Policy learning}
No learned-policy benefit is established here. Whether $g_t$ helps, in which
form, and whether it buys generalisation to unseen specifications, is the
central downstream experiment (Section~\ref{sec:policy}).

\paragraph{Embodiment transfer}
Sessions retarget onto a G1 in simulation. How much of a skilled human's
behaviour survives human-to-humanoid retargeting --- and whether what survives
is the part that matters for precision work --- is a question this pipeline
inherits from the retargeting literature rather than answers.

\paragraph{Physical deployment and safety}
Compatibility is shown in simulation. Physical execution raises separate
engineering and safety questions, particularly for a machine working near people
in a domain with a poor safety record.

\section{Conclusion}
\label{sec:conclusion}

Construction offers embodied AI something unusual: a machine-readable
description of a physical state before that state exists. Today's demonstration
datasets teach machines from what people saw and did. \system{} asks what becomes
possible when the record also contains what the person was trying to make true
--- registered to the workspace, visible to them while they worked, and
synchronised with how they moved.

The contribution is therefore not another capture pipeline but an interface for
collecting demonstrations in which physical state, intended state, and skilled
action are bound together at the moment of collection, and which remains
compatible with the humanoid infrastructure the field already has. One
instrumented session shows the binding surviving contact with a real task: a
specification that changes while the work proceeds, including when the worker
steps it back, recorded on the same clock as the body that responds to it, and
checkable against the physical scene from outside. Whether an
embodied learner can exploit that binding --- whether it can be told to build
something it has never built and do it --- remains open. That is precisely the
point: the data required to ask the question can now be collected.

\bibliographystyle{IEEEtran}
\bibliography{bib/references}

\end{document}